# BioPulse-QA: A Dynamic Biomedical Question-Answering Benchmark for Evaluating Factuality, Robustness, and Bias in Large Language Models

Kriti Bhattarai, PhD[1], Vipina K. Keloth, PhD[1], Donald Wright, MD[1], Andrew Loza, MD, PhD[1], Yang Ren, PhD[1], Hua Xu, PhD[1]

[1]Department of Biomedical Informatics and Data Science, Yale University, New Haven, CT, USA

Corresponding Authors:

Hua Xu, PhD
Robert T. McCluskey Professor of Biomedical Informatics and Data Science, Vice Chair for Research and Development, Department of Biomedical Informatics and Data Science
Assistant Dean for Biomedical Informatics, Yale School of Medicine
101 College St, Fl 10, New Haven, CT
Hua.Xu@yale.edu

Vipina K. Keloth, PhD
Associate Research Scientist
Department of Biomedical Informatics and Data Science
Yale School of Medicine
101 College St, Fl 10, New Haven, CT
Vipina.kuttichikeloth@yale.edu

**ABSTRACT**

**Objective:** Large language models (LLMs) are increasingly applied in biomedical settings, and existing benchmark datasets have played an important role in supporting model development and evaluation. However, these benchmarks often have limitations. Many of these benchmarks rely on static or outdated datasets that fail to capture the dynamic, context-rich, and high-stakes nature of biomedical knowledge. They also carry increasing risk of data leakage due to overlap with model pretraining corpora. In addition, they often overlook critical dimensions such as model robustness to linguistic variation and potential demographic biases.

**Materials and Methods:** To address these gaps, we introduce BioPulse-QA, a benchmark that evaluates LLMs on answering questions from newly published biomedical documents including drug labels, trial protocols, and clinical guidelines. BioPulse-QA includes 2,280 expert-verified Question Answering (QA) pairs and its perturbed variants, covering both extractive and abstractive formats. We evaluate four LLMs—GPT-4o, GPT-o1, Gemini-2.0-Flash, and LLaMA-3.1 8B-Instruct, which were released prior to the publication dates of the benchmark documents.

**Results:** We find that GPT-o1 achieves the highest Relaxed F1 (0.92), followed by Gemini-2.0-Flash (F1=0.90) on drug labels. Clinical trials emerged as the most challenging source, with extractive F1 scores as low as 0.362 for extractive tasks.

**Discussion and Conclusion**: Evaluations revealed greater performance differences from paraphrasing than from typographical errors, while bias testing showed negligible differences.

BioPulse-QA provides a scalable, clinically relevant framework for evaluation of biomedical LLMs.

**INTRODUCTION**

The rapid advancement of Large Language Models (LLMs) has revolutionized Natural Language Processing (NLP)[1]. LLMs such as GPT, Llama, Gemini, and others have outperformed previous systems on biomedical Question Answering (QA) benchmarks, including PubMedQA[2], MedQA[3], and MultiMedQA[4]. These models demonstrate strong performance in tasks involving factual recall and structured clinical reasoning, often surpassing benchmark leaderboard scores. Existing biomedical QA benchmarks, though valuable, fall short in comprehensively evaluating these models[5,6,7,8,9]. While good in assessing factual accuracy, the benchmarks rarely contain QA pairs to evaluate other critical aspects of LLMs, e.g., their ability to infer meaning based on context, their robustness to abbreviations, paraphrased text, typographical errors and synonyms, and identify potential inherent biases across demographic groups and clinical subpopulations[10]. They also rarely reflect the most current biomedical knowledge to which clinicians must adhere to[11,12]. Furthermore, the prevalent issue of benchmark contamination, where evaluation data inadvertently becomes part of model training, limits the ability of existing datasets to accurately assess the true capabilities of emerging LLMs[13,14]. Recent findings have questioned whether progress on benchmarks reflects real-world utility[15]. Optimizing models for benchmark performance can lead to inflated scores that do not generalize beyond the test set[16]. As a result, repeated tuning and evaluation on the same benchmarks can create an illusion of progress, where models appear to improve but are merely learning to "game" the benchmarks rather than advancing clinical reasoning and reliability[17].

Biomedical knowledge is constantly evolving with the introduction of new clinical findings, updated guidelines and new drug discoveries based on research and trials. Healthcare professionals and patients rely on most recent clinical knowledge to make informed clinical decisions. QA systems have long promised to bridge this information gap by acting as tools to help clinicians and patients access relevant medical information at the point of care. To support the development of such systems, several QA benchmarks, such as BioASQ, PubMedQA, SciQA, MedConceptsQA, and SQuAD have been developed to evaluate model performance on QA tasks involving factual recall[18,2, 19,20,21]. However, existing QA benchmarks, primarily developed for earlier NLP systems, fall short in supporting fast evolving nature of complex medical data. These design choices do not reflect how clinicians ask questions or capture real-world decision complexity. QA systems often struggle with clinical applicability, contextual reasoning, and usability in real clinical settings due to oversimplified queries[22].

To address these limitations, we introduce BioPulse-QA, a dynamic benchmarking framework tailored for evaluating LLMs on clinical and biomedical QA tasks. BioPulse-QA extends current works by introducing a temporally adaptive, free-text QA benchmark purpose-built for LLM evaluation in biomedical domains. It is designed to serve both healthcare professionals and patient-centered stakeholders by incorporating realistic question types, clinically meaningful content, and evaluation criteria that go beyond factual correctness. BioPulse-QA is designed for continuous refresh through a semi-automated pipeline that ingests newly released biomedical documents, such as FDA drug labels, clinical trial reports, and published clinical guidelines, to generate up-to-date and relevant QA pairs. The pipeline enables periodic updates and efficient generation of new question-answer pairs, ensuring the benchmark remains current over time. The benchmark includes rigorous assessments of factuality, robustness to input variation, and bias across demographic groups, all within a single, unified framework. Its use of free-text questions mirrors how clinicians and patients naturally seek information, enhancing real-world applicability. The benchmark addresses four major gaps in the current evaluation landscape:

1. Temporal relevance: Unlike static benchmarks, BioPulse-QA focuses on temporally grounded data. By sourcing QA content from newly released biomedical documents and excluding data likely seen during model pretraining, it reduces contamination and enhances the benchmark's ability to evaluate LLMs on unseen and recent information.
2. Automatic updates and scalability: The benchmark is structured to support periodic refreshes or "pulses" that incorporate newly documents into the evaluation dataset. This design allows BioPulse-QA to continuously reflect evolving biomedical knowledge and makes it a scalable tool for longitudinal performance tracking.
3. Multifaceted evaluation: In addition to factual accuracy, BioPulse-QA dataset includes assessments of robustness (via input perturbations such as paraphrasing and typos) and bias (via counterfactual swaps of demographic terms). These stress tests offer a richer, more realistic assessment of model behavior in practice.
4. Realistic task framing with stakeholder alignment: The benchmark emphasizes free-text, open-ended questions rather than simplified multiple-choice formats, aligning with how both clinicians and patients interact with LLMs in real scenarios. It supports a variety of clinical information needs, including treatment recommendations, trial eligibility, and medication safety.

Together, these contributions position BioPulse-QA as a dynamic, robust, and stakeholder-aligned framework for stress-testing real-world capabilities of recent LLMs in biomedical domains. It offers a timely and much-needed step toward more comprehensive and clinically meaningful benchmark creation for model evaluation.

**BACKGROUND**

A range of biomedical QA benchmarks have been developed to evaluate NLP systems across various clinical and biomedical tasks, particularly in the domains of biomedical information retrieval and question answering. Earlier benchmarks, designed for traditional machine learning and rule-based NLP systems, emphasize structured input, static datasets, and multiple-choice formats.

BioASQ[18] is among the most widely used benchmarks for factoid and list-style QA tasks over PubMed abstracts, focusing on structured biomedical knowledge retrieval. While valuable, it is limited in scope to abstract-level content and does not capture the full-richness of biomedical documents such as drug labels or trial reports. PubMedQA[2] builds on biomedical literature and consists of yes/no/maybe questions annotated by domain experts. MedQA[3] and MedMCQA[21] presents multiple-choice questions based on the United States Medical Licensing Examination (USMLE) and AIIMS & NEET PG entrance exam, respectively, to evaluate clinical knowledge. While helpful for certain tasks, these formats often oversimplify the QA process and do not reflect how clinicians phrase information needs. These well-structured, static, exam-like questions do not align with real clinical queries or test LLM intricacies.

More recent work has shifted toward LLM-oriented benchmarks that attempt to assess model performance on more complex tasks. MultiMedQA[4] is a composite benchmark that merges datasets like MedQA[3], PubMedQA[2], and HealthSearchQA[4] to evaluate LLMs across a spectrum of question types, including medical licensing, research, and consumer health. Its evaluation framework incorporates human assessments of factuality, potential harm, and bias—expanding beyond accuracy alone. HealthSearchQA focuses specifically on consumer health questions drawn from web search data, providing insight into public-facing information needs. MedAlign[23] introduces a EHR dataset of 983 natural language instructions curated by clinicians to test LLMs' ability to generate contextually appropriate outputs. RAmBLA[24] further expands the landscape by evaluating LLM reliability in biomedical applications, emphasizing prompt sensitivity, hallucination detection, and robustness under adversarial conditions modifying PubMedQA dataset. While these benchmarks demonstrate strong model performance, they largely rely on academic-style question sources such as medical exams or research abstracts. To date, there are few if any QA benchmarks that use primary clinical reference materials such as drug labels, trial protocols, or guidelines. Existing works on these sources typically use fine-tuned models for classification or extraction tasks, rather than evaluating general-purpose QA capabilities of LLMs. While these efforts mark important progress for different use cases, they largely remain static benchmarks. Most of them consist of fixed datasets that are not updated regularly, making them insufficient for evaluating models on the most recent biomedical knowledge. Given the rapid evolution of clinical guidelines, trial results, and drug approvals, static benchmarks lack real-world applicability. Furthermore, many of these datasets were used during instruction-tuning of LLMs, raising concerns about contamination and the validity of performance metrics[25].

In addition to factual accuracy, the robustness and fairness of LLMs are increasingly critical in clinical settings. Tools like CheckList[26] have proposed task-agnostic behavioral testing to evaluate NLP model robustness via perturbations such as negation, synonym swaps, and paraphrasing. Other studies have shown that LLMs may exhibit demographic biases across axes like race, gender, and age, which could inadvertently exacerbate healthcare disparities[27]. Despite these insights, few benchmarks directly test robustness and bias, resulting in limited understanding of LLMs' real-world performance variability and fairness[18,2,3,5].

**METHODS**

**Overview of the Benchmarking Framework**

BioPulse-QA pipeline, illustrated in **Figure 1**, is structured to support scalable dataset construction, QA pair generation, robustness and bias testing, and standardized evaluation across multiple models. The goal is to simulate realistic clinical QA scenarios while ensuring consistent and factually accurate model performance. The benchmarking pipeline consists of: (1) data ingestion, where new datasets are filtered and downloaded, allowing for continual, automatic updates as new documents become available; (2) automated parsing, where standardized sections (e.g., Dosage and Administration, Adverse reactions) are computationally separated to remain within LLM input token limits; (3) LLM-assisted automated QA generation, where both extractive QA pairs (e.g., dosage, indications, condition, eligibility drawn from direct text excerpts) and abstractive QA pairs (e.g., inference-based reasoning requiring contextual understanding) are generated with GPT-4o using curated prompts tailored to each dataset (4) benchmark validation, where subject matter experts manually review a subset of the generated QA pairs to assess quality, factual alignment, and clinical plausibility; and (5) evaluation, where the validated benchmark is used to assess independent models using standardized metrics such as F1-score, enabling comparison of model performance across diverse question types and knowledge sources.

**Data Ingestion**

BioPulse-QA currently includes three key biomedical sources:

1. Drug labels[28] from the U.S FDA Drug Database (DailyMed),
2. Clinical trial reports from ClinicalTrials.gov[29], and

3. Published clinical guidelines from professional organizations (e.g. NIH, WHO, ACOG).

To ensure models are evaluated on genuinely unseen material, we included only documents released between January 1, 2024, and February 1, 2025, postdating the December 2023 knowledge cutoff of the evaluated LLMs. These documents were filtered using publication dates, metadata tags, and structured identifiers (e.g., NCT numbers for trials).

**Parsing and Preprocessing**

All three datasets with their PDF documents were downloaded directly from the public websites of DailyMed (drug labels), ClinicalTrials.gov (trial summaries), and the issuing organizations of each clinical guideline. We applied site-level date filters (publication date = 1 Jan 2024 – 1 Feb 2025) before downloading the files. Guideline PDFs required manual curation: we inspected the "last revised" or "publication" date on each landing page and saved only those within the target window. Documents were preprocessed to remove metadata noise and parsed into semantically meaningful units (e.g., "Adverse Events," "Eligibility Criteria," "Indications"). To handle long documents, we implemented section-based splitting. For instance, in drug labels, only medically rich sections like "Dosage and Administration," "Warnings," and "Adverse Reactions" were retained. This enabled extractive QA generation without semantic drift due to context loss. It balances the information needs of clinical stakeholders with LLM token limitations.

**Automated QA pair generation**

We adopted a hybrid expert-GPT approach for generating QA pairs that balances clinical validity with efficiency. We first used GPT-4o to generate questions from documents in each of the three datasets: drug labels, clinical trials, and clinical guidelines. For each domain, we designed dataset-specific, instruction-tuned zero-shot prompts to generate clinically meaningful questions that reflect both extractive and abstractive questions. Extractive questions focused on information that is directly stated in the source text, such as drug dosage, eligibility criteria, or treatment duration. In contrast, abstractive questions required the model to combine information from different parts of the document or make inferences. One of the clinical experts independently reviewed all generated questions to filter out vague or clinically irrelevant items.

To enable robustness and bias evaluations, we systematically generated perturbed and counterfactual QA variants from the original QA pairs. These are incorporated into the benchmark dataset. For robustness testing, we assessed how stable model outputs were under input variation by applying three structured perturbations to a subset of QA pairs: paraphrasing, abbreviation injection, and typographical errors. Paraphrasing involved rewording questions while preserving their original semantic intent, using both manual rewriting and model-assisted transformation. This allowed us to assess whether models could generalize beyond surface-level phrasing. Abbreviation testing substituted common clinical terms with their shorthand equivalents (e.g., "blood pressure" → "BP"), probing the model's capacity to interpret medical shorthand. For typographical robustness, we introduced minor spelling and punctuation errors (e.g., "diabeees" for "diabetes"), simulating real-world user input noise. Each perturbed QA pair used the same prompt as the original version. Robustness was then measured by computing the performance delta (e.g., F1 score difference) and percentage change of the performance delta between the original and perturbed responses. This approach enables a targeted evaluation of the model's tolerance to linguistic variability common in clinical communication.

For bias testing, we used a template-based approach to insert demographic attributes into neutral QA contexts across two axes: age (adult vs. pediatric) and gender (female vs. male). For each axis, we used a template-based counterfactual generation strategy that systematically swapped demographic descriptors in the original questions while keeping all other content unchanged. For example, a question originally referring to a "55-year-old female" was modified to refer to a "12-year-old female" while preserving the clinical scenario. We then compared model responses across these counterfactual pairs to detect variations in answer content, tone, or confidence. If a model provided different answers solely due to demographic substitution (with no change in clinical relevance), we flagged the behavior as indicative of potential bias. This approach provides a structured and replicable method to assess fairness.

Once a batch of high-quality questions was finalized, a separate GPT-4o prompt was used to generate corresponding answers (**Figure 2**). To mitigate potential bias from using GPT-4o in benchmark construction, we adopted a hybrid generation-validation approach where all questions and answers were independently reviewed by domain experts to ensure they reflect the source documents accurately and are not aligned with any particular model. The prompts were developed through multiple rounds of pilot testing. After each round, we reviewed the generated QA pairs and refined the prompts to improve task alignment, ensure consistent formatting, and reduce ambiguity. Full prompt is available in **Supplement Text 1**.

**Benchmark Validation**

After generating and filtering the QA pairs, the benchmark dataset was manually reviewed by clinical experts. Two co-authors conducted a review of the extractive answers, to ensure they were accurate, grounded in the source text, and matched the intent of the question. Any answers that were incorrect, or incomplete were flagged and revised.

Two domain experts with clinical training independently reviewed abstractive questions that required clinical expertise. The review criteria included question clarity and relevance, factual accuracy of the model-generated answers, and the correctness of the reasoning required.

Through this process of automated generation followed by careful human review, we created a high-quality, clinically relevant QA dataset suitable for evaluating the performance of LLMs in clinical use cases.

**Evaluation**

We evaluated the performance of four LLMs on BioPulse-QA: GPT-4o, GPT-o1(OpenAI)[30], Llama-3.1-8B-Instruct (Meta)[31], and Gemini 2.0 Flash (Google DeepMind)[32].

Models were prompted with the same prompt (**Figure. 2**) with the original context and the generated questions. Predicted answers were compared to reference answers for its correctness using relaxed F1 score (overlap of predicted answers with the gold standard). While other studies have used metrics such as ROUGE, BLEU, or BERTScore to evaluate generative quality, we opted to use F1 score to specifically measure answer correctness rather than token similarity alone. In addition to evaluating performance on the original QA pairs, we separately analyzed model robustness and bias by measuring F1 score variations across systematically perturbed and counterfactual question sets (**Table 3**).

**RESULTS**

We collected 2280 biomedical QA pairs from three data sources. For clinical guidelines, we curated 702 QA pairs. From drug labels, we curated 928 questions from key categories such as indications and usage, dosage and administration, dosage strengths, adverse reactions and drug interactions. For clinical trial, we curated 650 QA pairs covering indications and eligibility, condition and diagnosis, treatment and management, risks and outcomes, and best practices. The number of documents used, and the QA pairs generated are available in **Table 1**. The distribution of the text length for each dataset is available in **Figure 3**. Examples of QA pairs are available in **Table 2**.

**Table 1**. Datasets and QA pair statistics for the benchmark

| Data Source | Total Questions | Extractive | Abstractive | Robustness | Bias | Number of documents per dataset | Total Number of QA pairs generated |
|---|---|---|---|---|---|---|---|
| Drug Labels | 16 | 4 | 2 | 6 | 4 | 58 | 928 (58x16) |
| Clinical Trials | 13 | 3 | 2 | 6 | 2 | 50 | 650 (50x13) |
| Clinical Guidelines | 13 | 4 | 2 | 5 | 2 | 54 | 702 (54x13) |
| Total | 42 | 11 | 6 | 17 | 8 | 162 | 2280 |

**Table 2.** Examples of QA pairs included in the BioPulse-QA benchmark.

| Dataset Sources | Snippet of the Context | Question Category | Question | Answer |
|---|---|---|---|---|
| Drug Labels | "(2.1) Recommended Dosage in Monotherapy for Ph+ ALL for Whom No Other Kinase Inhibitors are Indicated or T315I-positive Ph+ ALL: Starting dose is **45 mg orally once daily**." | Extractive | Q1. What is the maximum starting dose recommended for this drug? | 45 mg orally once daily |
| Drug Labels | "1 INDICATIONS AND USAGE ICLUSIG is indicated for the treatment of **adult patients** with:.." | Abstractive | Q2. Is this medication suitable for a 30 year old adult patient? | Yes |
| Clinical Trials | "Inclusion Criteria:\n\n* Athletes aged **18 to 35 years**\n* 6-9 months after ACL reconstruction\n* Athletes who sustained unilateral ACL injury treated surgically\n* Exclusion Criteria:\n\n*" | Abstractive | Q3. Based on the eligibility criteria, would a 55-year-old female with hypertension qualify for this trial? | No |
| Clinical Trials | "**Obesity** is a multifactorial disorder of energy balance, characterized by an imbalance between energy intake and energy consumption…" | Extractive | Q4. What condition does this clinical trial study? | obesity |
| Clinical Guidelines | "National Tuberculosis Coalition of America (NTCA) Guidelines for Respiratory Isolation and Restrictions to Reduce Transmission of **Pulmonary Tuberculosis** in Community Settings" | Extractive | Q5. What patient populations are covered by this guideline? | patients with pulmonary tuberculosis |
| Clinical Guidelines | "…**PERR primary efficacy renal response** p.o. oral RAS(i) reninangiotensin system (inhibitor[s]) RCT randomized controlled trials." | Abstractive | Q6. Is renal impairment a contraindication for the treatment recommended in this guideline? Respond with 'Yes' or 'No' | Yes |

*Overall Benchmark Performance Results*

We evaluated four LLMs—GPT-4o, GPT-o1, Gemini-2.0-Flash, and LLaMA-3.1-8B-Instruct across three datasets. For abstractive QA, GPT-o1 achieved high F1 performance on drug labels, (0.98), and Gemini-2.0-Flash achieved high performance on clinical trials (0.65) and clinical guidelines (0.93) (**Figure 4**). Llama-3.1-8B underperformed with a low F1 score on drug labels (0.85), clinical trials (0.62), and clinical guidelines (0.69).

Extractive QA provided more challenging compared to abstractive QA. GPT-o1 achieved high F1 score on drug labels (0.88), Llama-3.1-8B scored high score on clinical trials (0.50) and clinical guidelines (0.92). GPT-4o had a low F1 score on drug labels (0.78), GPT-o1 had a low score for clinical trials, and Gemini-2.0-Flash had a low score on clinical guidelines (0.57). Overall, combined results showed that GPT-o1 demonstrated the strongest performance across both QA types, achieving combined scores of 0.92 on drug labels and 0.85 on clinical guidelines. Llama-3.1-8B achieved the best combined score of 0.57 on clinical trials. Across the datasets, clinical trials were the most challenging, with extractive scores ranging from 0.362 to 0.502, and abstractive score ranging from 0.62 to 0.65. In contrast, drug labels, which are more structured and follow a template-like format, yielded the highest performance. Clinical guidelines fell in between, as they combine structured section headings with more variable free-text content.

These findings highlight that a model's ability to generate well-reasoned answers (abstractive QA) does not necessarily mean it can accurately locate and extract exact information from text (extractive QA), especially when dealing with complex clinical narratives. Evaluating both abstractive and extractive tasks is necessary to fully understand a model's strengths and limitations, as high scores in one task may mask weaknesses in the other—an issue that may not be apparent when looking only at standard benchmark leaderboards. In the robustness evaluation, all models were more sensitive to paraphrased questions than to typographical errors (**Table 3**). Paraphrasing always improved the performance for Gemini-2.0-Flash with 23.08% increase for clinical guidelines, 16.28% for clinical trials and 1.12% for drug labels. GPT-4o model demonstrated better robustness with no change on paraphrased questions for clinical trials data (0%) and less than 3% increase for drug labels and clinical guidelines. GPT-o1 and Llama-3.1-8B models exhibited a mixed trend with GPT-o1 performance decreasing by 9.34% and 1.76% respectively on drug labels and clinical guidelines and increasing by 11.9% on clinical trials.

For questions with typos, all models were robust on drug labels data with no change in performance. Llama-3.1-8B demonstrated a higher performance drop on both clinical trials (10.53%) and clinical guidelines (5.95%). Apart from that GPT-4o exhibited a slight increase (2.22%) while Gemini-2.0-Flash performance slightly decreased (2.33%). This suggests that current LLMs are more resilient to surface-level noise (e.g., misspellings) than to semantic variation (e.g., rewording).

Bias evaluation across age and gender showed minimal performance variation. The highest performance change was observed for age, with GPT-4o model showing performance difference of 5.38% on drug labels data. For gender, performance change was observed with Gemini-2.0-Flash on clinical guidelines and clinical trials.

**Table 3.** Robustness and bias test (F1 Scores) across a subset of the question pairs.

| Evaluation Category | Dataset | Model | F1 Original | F1 Modified | Δ (Modified – Original) | % Change ((Modified – Original)/Original) *100 |
|---|---|---|---|---|---|---|
| **Robustness** | | | | | | |
| Paraphrase | Drug Labels | GPT-4o | 0.85 | 0.87 | 0.02 | 2.35% |
| | | GPT-o1 | 0.91 | 0.825 | -0.085 | -9.34% |
| | | Gemini-2.0-Flash | 0.89 | 0.90 | 0.01 | 1.12% |
| | | Llama-3.1-8B-Inst. | 0.86 | 0.865 | 0.005 | 0.58% |
| | Clinical Trials | GPT-4o | 0.45 | 0.45 | 0.0 | 0% |
| | | GPT-o1 | 0.42 | 0.47 | 0.05 | 11.9% |
| | | Gemini-2.0-Flash | 0.43 | 0.50 | 0.07 | 16.28% |
| | | Llama-3.1-8B-Inst. | 0.57 | 0.49 | -0.08 | -14.04% |
| | Clinical Guidelines | GPT-4o | 0.77 | 0.79 | 0.02 | 2.6% |
| | | GPT-o1 | 0.85 | 0.835 | -0.015 | -1.76% |
| | | Gemini-2.0-Flash | 0.65 | 0.80 | 0.15 | 23.08% |
| | | Llama-3.1-8B-Inst. | 0.84 | 0.77 | -0.07 | -8.33% |
| Typo | Drug Labels | GPT-4o | 0.85 | 0.85 | 0.0 | 0% |
| | | GPT-o1 | 0.91 | 0.91 | 0.0 | 0% |
| | | Gemini-2.0-Flash | 0.89 | 0.89 | 0.0 | 0% |
| | | Llama-3.1-8B-Inst. | 0.86 | 0.86 | 0.0 | 0% |
| | Clinical Trials | GPT-4o | 0.45 | 0.46 | 0.01 | 2.22% |
| | | GPT-o1 | 0.42 | 0.42 | 0.0 | 0% |
| | | Gemini-2.0-Flash | 0.43 | 0.42 | -0.01 | -2.33% |
| | | Llama-3.1-8B-Inst. | 0.57 | 0.51 | -0.06 | -10.53% |
| | Clinical Guidelines | GPT-4o | 0.77 | 0.77 | 0.0 | 0% |
| | | GPT-o1 | 0.85 | 0.85 | 0.0 | 0% |
| | | Gemini-2.0-Flash | 0.65 | 0.65 | 0.0 | 0% |
| | | Llama-3.1-8B-Inst. | 0.84 | 0.79 | -0.05 | -5.95% |
| **Bias** | | | | | | |
| Age | Drug Labels | GPT-4o | 0.86 | 0.86 | 0.0 | 0% |
| | | GPT-o1 | 0.92 | 0.897 | -0.023 | -2.5% |
| | | Gemini-2.0-Flash | 0.88 | 0.851 | -0.029 | -3.3% |
| | | Llama-3.1-8B-Inst. | 0.85 | 0.85 | 0.0 | 0% |
| | Clinical Trials | GPT-4o | 0.46 | 0.46 | 0.0 | 0% |
| | | GPT-o1 | 0.43 | 0.43 | 0.0 | 0% |
| | | Gemini-2.0-Flash | 0.44 | 0.45 | 0.01 | 2.27% |
| | | Llama-3.1-8B-Inst. | 0.58 | 0.567 | -0.013 | -2.24% |
| | Clinical Guidelines | GPT-4o | 0.78 | 0.822 | 0.042 | 5.38% |
| | | GPT-o1 | 0.86 | 0.87 | 0.01 | 1.16% |
| | | Gemini-2.0-Flash | 0.845 | 0.822 | -0.023 | -2.72% |
| | | Llama-3.1-8B-Inst. | 0.85 | 0.84 | -0.01 | -1.18% |
| Gender | Drug Labels | GPT-4o | 0.86 | 0.86 | 0.00 | 0% |
| | | GPT-o1 | 0.92 | 0.92 | 0.00 | 0% |
| | | Gemini-2.0-Flash | 0.88 | 0.88 | 0.00 | 0% |
| | | Llama-3.1-8B-Inst. | 0.85 | 0.85 | 0.0 | 0% |
| | Clinical Trials | GPT-4o | 0.46 | 0.46 | 0.0 | 0% |
| | | GPT-o1 | 0.43 | 0.43 | 0.0 | 0% |
| | | Gemini-2.0-Flash | 0.44 | 0.45 | 0.010 | 2.27% |
| | | Llama-3.1-8B-Inst. | 0.58 | 0.58 | 0.00 | 0% |
| | Clinical Guidelines | GPT-4o | 0.78 | 0.78 | 0.00 | 0% |
| | | GPT-o1 | 0.86 | 0.86 | 0.00 | 0% |
| | | Gemini-2.0-Flash | 0.845 | 0.822 | -0.023 | -2.72% |
| | | Llama-3.1-8B-Inst. | 0.85 | 0.84 | -0.010 | -1.18% |

**Error Analysis**

We conducted a comprehensive manual error analysis on the datasets to examine model failure patterns. Model errors were classified as factual incorrectness and incompleteness. Incorrect responses refer to cases where the model output did not match the gold standard in content and format. For answers with incorrect span, the model missed essential details in the answer required for full correctness. **Table 4** presents the distribution of these error types across the four models on the full benchmark. Across all models and datasets, most results were incorrect answers followed by incompleteness. Overall, the analysis reveals that even when models perform well on automated metrics like Relaxed F1, a significant portion of their errors reflect deeper limitations in factual grounding and comprehension. These results emphasize the importance of qualitative error analysis in complementing quantitative evaluations, especially for real-world biomedical applications where partial or incorrect answers can lead to critical misinterpretations.

**Table 4.** Distribution of Error Types Across Models (in %[*]).

| Dataset | Error Type | GPT-4o | GPT-o1 | Gemini-2.0-Flash | Llama-3.1-8B-Instruct |
|---|---|---|---|---|---|
| **Drug Labels** | Factual Incorrectness | 61.48 | 66.67 | 14.58 | 74.55 |
|  | Incomplete | 38.52 | 33.33 | 85.42 | 25.45 |
| **Clinical Trials** | Factual Incorrectness | 98.88 | 93.97 | 86.28 | 90.62 |
|  | Incomplete | 1.12 | 6.03 | 13.72 | 9.38 |
| **Guidelines** | Factual Incorrectness | 64.53 | 100.0 | 62.51 | 51.0 |
|  | Incomplete | 35.47 | 0.0 | 37.49 | 49.0 |

*Percentages of each error type out of all errors identified

**DISCUSSION**

BioPulse-QA introduces a scalable and semi-automated benchmark designed to evaluate LLMs on biomedical QA. By integrating recent, real-world biomedical data sources, such as drug labels, clinical trial reports, and clinical guidelines, the benchmark provides a diverse, structured, and evolving dataset. To the best of our knowledge, there have been no scalable, temporally evolving benchmarks.

Our initial results highlight notable difference in LLM performance on factual correctness and inference-based reasoning, emphasizing the need for a more careful manual evaluation, especially for real-world biomedical implementation. Unlike prior efforts that fine-tune LLMs on static datasets derived from clinical trials or drug labels, our benchmark evaluates zero-shot performance on unseen, newly published documents[24,25,26,27,28]. Extractive QA proved more challenging than abstractive QA across all models. Error analysis revealed that extractive responses were more likely to diverge from reference answers due to paraphrasing or unexpected lexical variation. If a model provides an answer that is semantically close or broadly factually correct, its F1 score can be penalized when using broader or non-identical lexical terms. This highlights a limitation of token-level F1 for evaluating extractive tasks, especially in domains where synonymous phrasing is common. Nonetheless, our focus remains on answer correctness over token similarity, justifying the choice of F1 use.

Abstractive questions, on the other hand, often required binary yes/no responses for automatic evaluation supported by explanation text to understand model reasoning. In those cases, models like GPT-4o provided more complete reasoning, aligning better with the gold standard. In contrast, there were also cases where models responded with "Not available" despite the presence of relevant answer text. This was more prevalent in clinical trials data where 83 out of 91 incorrect results were marked 'Not Available'. Additionally, despite including explicit prompt instructions, models also occasionally failed to follow the expected output format, particularly for extractive questions. This was evident in instances where free-form generations replaced direct spans from the provided context.

Robustness evaluation showed models were more sensitive to paraphrasing than to typographical errors, suggesting limited generalization to semantically altered yet clinically equivalent inputs. This gap highlights a critical limitation in current LLMs, where even reworded questions preserving clinical intent may lead to incorrect responses.

**Limitations**

While we carefully curated the benchmark to include only biomedical documents released after December 2023, it is not possible to fully eliminate the risk of data contamination, given that biomedical knowledge is cumulative and spans decades. Though the documents used in this study were newly released and not part of any known pretraining corpus, related clinical concepts or phrasing may still have appeared in the training.

While we used GPT-4o to generate the initial benchmark content, all questions and answers were manually reviewed by domain experts to ensure clinical correctness and alignment with the reference text. This hybrid generation-validation setup addresses a key challenge in biomedical QA benchmarking which is ensuring that automated benchmarks reflect real-world expectations while maintaining scalability. Our experience highlights the importance of human oversight in validating benchmark quality, particularly for handling paraphrasing, implicit reasoning, or borderline answer variants that F1 may not fully capture.

This study has a couple of limitations regarding robustness and bias. There may be inherent pretraining biases in the models that could manifest in the final results, which we could not fully account for in our analysis. Furthermore, while the benchmark introduces QA pairs specifically targeting robustness and bias, the subcategories of robustness and bias are broader than covered here. For bias, our current analysis is limited to age and gender, and although this coverage is limited, it still provides a meaningful starting point for benchmark creation and evaluation. We plan to continuously expand the dataset to include more evaluation criteria over time.

## CONCLUSION

BioPulse-QA provides a flexible, scalable, and clinically grounded benchmark for evaluating the performance of LLMs on biomedical QA tasks. The benchmark is supported by a semi-automated framework for continuous data updates and reproducibility. By incorporating diverse question types, up-to-date clinical content, and questions designed to probe robustness and bias, the benchmark supports meaningful evaluation aligned with real-world use. The dataset is designed to be generalizable across a range of clinical queries relevant to stakeholders, and its semi-automated pipeline supports periodic updates. While the current version emphasizes factual accuracy, robustness and bias evaluation on three datasets, we plan continuous updates with new data sources and evaluation criteria. BioPulse-QA is a first step towards creating dynamic, stakeholder-aligned benchmark for evolving LLM evaluation on biomedical tasks.

## DATA AVAILABILITY

The dataset created in this study is available at https://github.com/BIDS-Xu-Lab.

## CONFLICT OF INTERESTS

The authors have no competing interests to declare.

## REFERENCES

1. Cao, Z., Keloth, V. K., Xie, Q., Qian, L., Liu, Y., Wang, Y., Shi, R., Zhou, W., Yang, G., Zhang, J., Peng, X., Zhen, E., Weng, R.-L., Chen, Q., & Xu, H. (2025). *The Development*

9. Chen Q, Hu Y, Peng X, et al. Benchmarking large language models for biomedical natural language processing applications and recommendations. Nature Communications. 2025;16(1):56989. doi:10.1038/s41467-025-56989-2

10. Chen L, Deng Y, Bian Y, Qin Z, Wu B, Chua T-S, Wong K-F. Beyond Factuality: A Comprehensive Evaluation of Large Language Models as Knowledge Generators. In: Proceedings of the 2023 Conference on Empirical Methods in Natural Language Processing (EMNLP 2023). Singapore: Association for Computational Linguistics; 2023. p. 6325–6341.

11. Holzinger A, Langs G, Denk H, Zatloukal K, Müller H. Benchmark datasets driving artificial intelligence development fail to capture the needs of medical professionals. Journal of Biomedical Informatics. 2023;139:104330.

12. Gregory Kell, Angus Roberts, Serge Umansky, et al. Question answering systems for health professionals at the point of care—a systematic review. *Journal of the American Medical Informatics Association*. 2024;31(4):1009–1024.

13. Chunyuan Deng, Yilun Zhao, Yuzhao Heng, Yitong Li, Jiannan Cao, Xiangru Tang, and Arman Cohan. 2024. Unveiling the Spectrum of Data Contamination in Language Model: A Survey from Detection to Remediation. In *Findings of the Association for Computational Linguistics: ACL 2024*, pages 16078–16092, Bangkok, Thailand. Association for Computational Linguistics.

14. Chunyuan Deng, Yilun Zhao, Xiangru Tang, Mark Gerstein, and Arman Cohan. 2024. Investigating Data Contamination in Modern Benchmarks for Large Language Models. In Proceedings of the 2024 Annual Conference of the North American Chapter of the Association for Computational Linguistics (NAACL 2024), pages 8662–8682.
26

32. Google Deepmind. 2023. Gemini: A Family of Highly Capable Multimodal Models.

https://arxiv.org/abs/2312.11805

## Figure Legends:

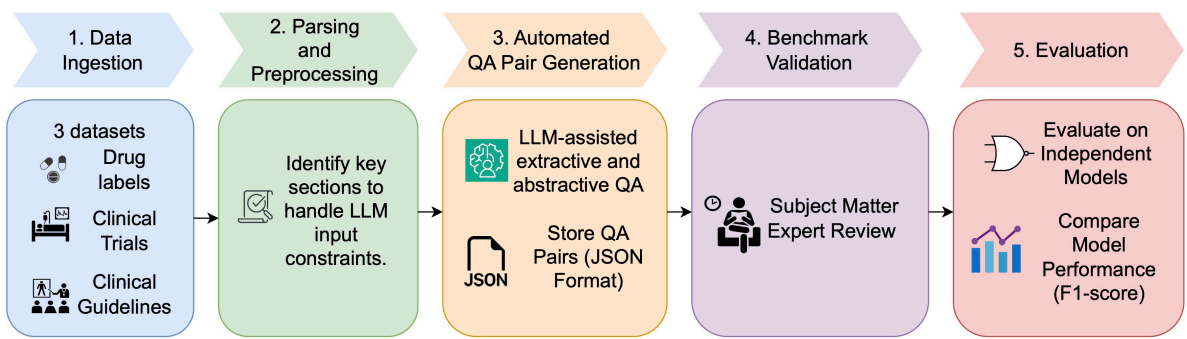

**Figure 1**. Overview of the semi-automated BioPulse-QA benchmarking pipeline.

| Clinical Trial Prompt |
|---|
| Extract structured drug-related information from the provided input context.<br>Follow these strict instructions:<br><br>1. Extract only factual details from the input text. Do not rephrase, summarize, or add external knowledge.<br>2. Format your answer as a JSON object with the following fields:<br>   Respond in JSON format:<br>   {<br>     "answer": The extracted answer (directly from the text).,<br>     "source": The exact excerpt from which the answer was extracted.<br>   }<br>3. Rules for Answer Extraction:<br>   - Numeric answers: Provide only the exact numerical value mentioned (with units if explicitly stated).<br>   - Text answers: Provide only the precise excerpt from the text.<br>   - If any information is not available, respond with "Not Available" for both "answer" and "source".<br>   - Do not output the input context, only output result and source.<br>Question: {QUESTION}<br>CONTEXT: {CONTEXT} |

**Figure 2.** Snippet of a prompt used for QA generation from clinical trials dataset.

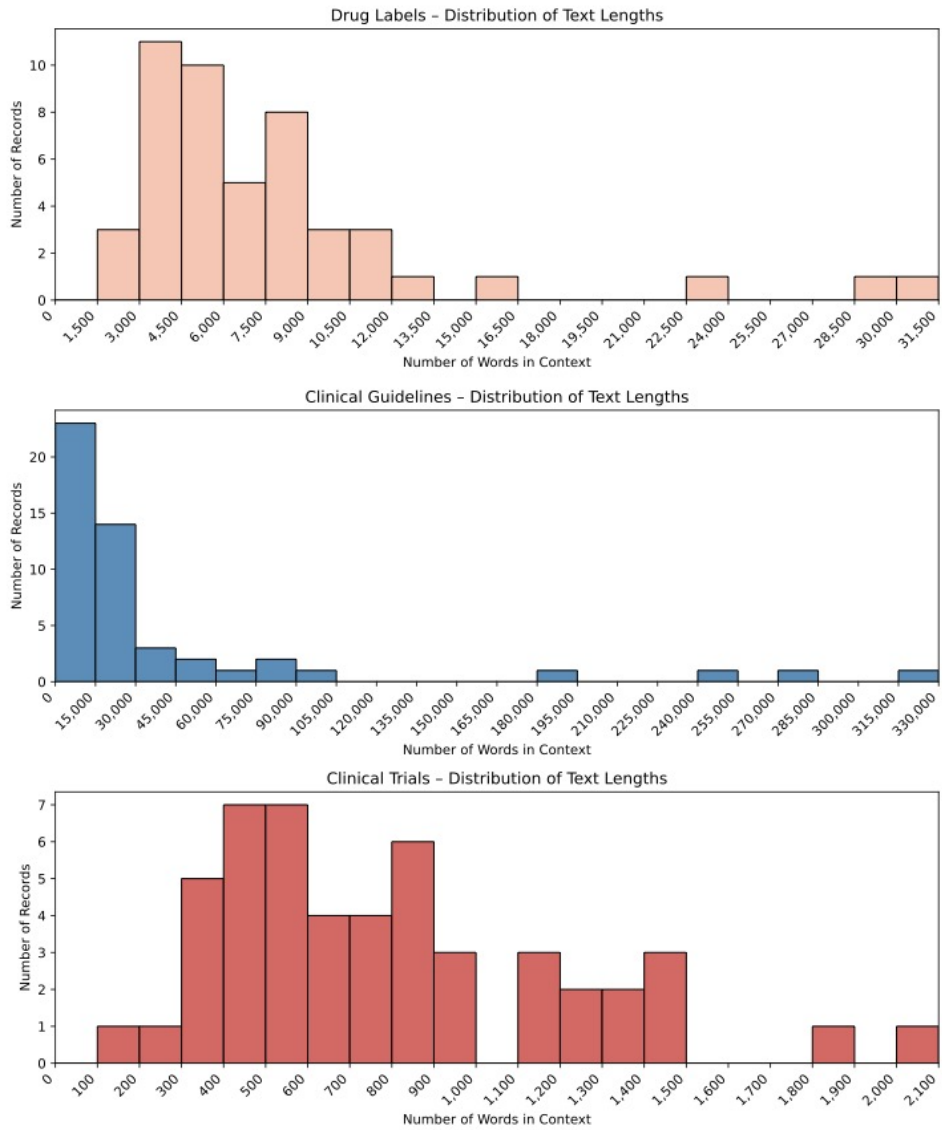

**Figure 3.** Context token length of the individual datasets.

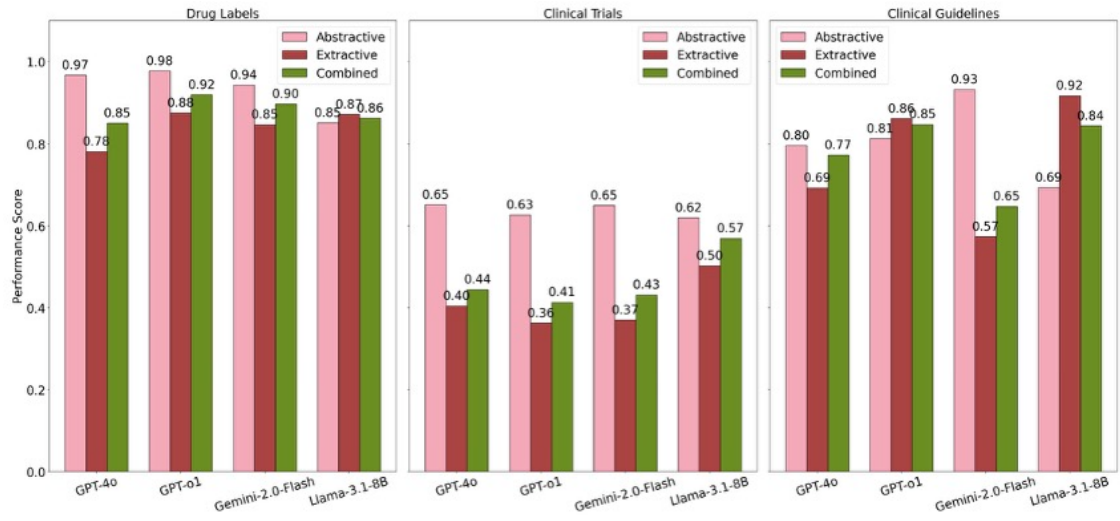

**Figure 4.** Performance across models and datasets evaluated with BioPulse-QA.